\documentclass{article}

\usepackage{microtype}
\usepackage{graphicx}
\graphicspath{ {./img/} }
\usepackage{subfigure}
\usepackage{booktabs} 
\usepackage{bm}

\usepackage{hyperref}

\usepackage[accepted]{icml2021}

\icmltitlerunning{A Closer Look at Cross-modal Transfer of Pretrained Transformers}

\begin{document}
\newcounter{commentCounter}
\newif\iftrvar
\trvartrue
\iftrvar
\newcommand{\tim}[1]{{\small \color{red}
\refstepcounter{commentCounter}\textsf{[TR]$_{\arabic{commentCounter}}$:{#1}}}}
\def\todo{\textbf{\color{red}TODO}}
\newcommand{\incharge}[1]{\textcolor{red}{\textbf{\MakeUppercase{#1}}}}
\newcommand{\todoc}[1]{{\color{red}\textbf{TODO} #1}}
\def\toref{[\textbf{\color{purple}REF}]}
\newcommand{\torefc}[1]{[\textbf{\color{purple}REF} {\color{purple} #1}]}
\newcommand{\danielle}[1]{{\small \color{blue}\refstepcounter{commentCounter}\textsf{[D]$_{\arabic{commentCounter}}$:{#1}}}}
\else
\newcommand{\tim}[1]{}
\def\todo{}
\def\todoc{}
\def\toref{}
\def\torefc{}
\fi

\def\blah{\textbf{\color{red}???}}

\definecolor{lightgray}{rgb}{.9,.9,.9}
\definecolor{darkgray}{rgb}{.4,.4,.4}
\definecolor{purple}{rgb}{0.65, 0.12, 0.82}
\definecolor{darkgreen}{rgb}{0, 0.365, 0}

\twocolumn[

\icmltitle{Don't Sweep your Learning Rate under the Rug:\\A Closer Look at Cross-modal Transfer of Pretrained Transformers}



\icmlsetsymbol{equal}{*}

\begin{icmlauthorlist}
\icmlauthor{Danielle Rothermel}{fb}
\icmlauthor{Margaret Li}{uw}
\icmlauthor{Tim Rockt\"{a}schel}{fb,uc}
\icmlauthor{Jakob Foerster}{fb}
\end{icmlauthorlist}

\icmlaffiliation{fb}{Facebook AI Research}
\icmlaffiliation{uw}{University of Washington}
\icmlaffiliation{uc}{University College London}

\icmlcorrespondingauthor{Danielle Rothermel}{drotherm@fb.com}

\icmlkeywords{Machine Learning, ICML}

\vskip 0.3in
]



\printAffiliationsAndNotice{}  

\begin{abstract}
Self-supervised pre-training of large-scale transformer models on text corpora followed by finetuning has achieved state-of-the-art on a number of natural language processing tasks. Recently, \citet{Lu2021pretrained} claimed that \emph{frozen} pretrained transformers (FPTs) match or outperform training from scratch as well as unfrozen (fine-tuned) pretrained transformers in a set of transfer tasks to \emph{other modalities}. In our work, we find that this result is, in fact, an artefact of not tuning the learning rates. After carefully redesigning the empirical setup, we find that when tuning learning rates properly, pretrained transformers do outperform or match training from scratch in all of our tasks, but only as long as the \emph{entire model} is fine-tuned. Thus, while transfer from pre-trained language models to other modalities does indeed provide gains and hints at exciting possibilities for future work, properly tuning hyperparameters is important for arriving at robust findings.
\end{abstract}

\begin{figure}[t]
    \centerline{\includegraphics[width=7cm]{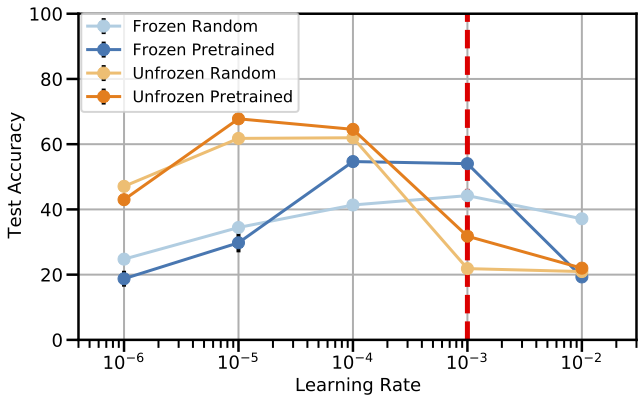}}
    \caption{Test accuracy on the CIFAR10 LRA task across the learning rate sweep, with error bounds across 3 seeds.  The learning rate reported by \citet{Lu2021pretrained}, $1 \times 10^{-3}$, is marked with a dashed red line, demonstrating that any lower learning rate would have given inverted results on this task.}
    \label{fig:lrsweep_cifar10_lra_annotated}
\end{figure}

\begin{figure*}[t]
	\vspace*{1.5em}
    \centerline{\includegraphics[width=14cm]{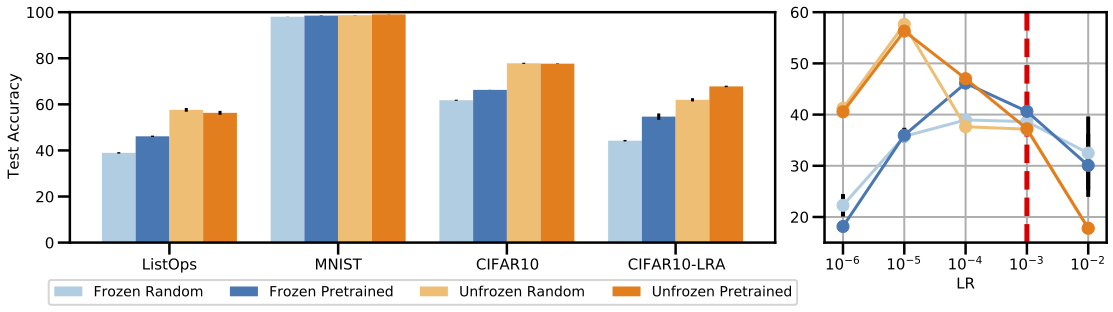}}
    \caption{[Left] Test accuracy of the best learning rate for all settings and tasks, with error bounds over 3 seeds. Exact values reported in Table \ref{main-table}. The Frozen variants under-perform across all tasks, but the Unfrozen Pretrained variant matches or exceeds all other variants. [Right] The test accuracy on the ListOps task across the learning rate sweep, error bounds over 3 seeds. The learning rate reported by \citet{Lu2021pretrained}, $1 \times 10^{-3}$, is marked with a dashed red line, demonstrating that any lower learning rate would have given inverted results on this task. Each learning rate evaluated between $1\times 10^{-5}$ and $1 \times 10^{-3}$ leads to a different conclusion about the best architecture.}
    \label{fig:mainresults}
\end{figure*}

\section{Introduction}
\label{introduction}

Transformer-based pretrained language models (LMs) have led to a revolution in the area of natural language processing (NLP) in recent years~\cite{10.5555/3295222.3295349}, a progress fueled by larger training data sets, bigger models and increasing computational demand.
Finetuning such pretrained LMs has led to state-of-the-art performance across a variety of NLP benchmarks~\cite{petroni-etal-2021-kilt,wang-etal-2018-glue,NEURIPS2019_4496bf24}. 
In these settings, the LM is pretrained on a large collection of natural language texts such as the Google Book Corpus~\cite{10.1109/ICCV.2015.11} or the 1B Word Benchmark \cite{chelba2014one}. Subsequently, the model is fine-tuned on a given task of interest, e.g. sentiment analysis ~\cite{maas-etal-2011-learning} or text classification~\cite{NIPS2015_250cf8b5}. 

While the ability to transfer language representations, e.g. word representations~\cite{NIPS2013_9aa42b31, pennington-etal-2014-glove} or contextual representations~\cite{devlin-etal-2019-bert,radford2019language}, between different language tasks has been well studied and revealed few-shot~\cite{NEURIPS2020_1457c0d6} and zero-shot~\cite{petroni-etal-2019-language} abilities, recent work has focused on the exciting possibility of transfer between different modalities~\cite{Lu2021pretrained}.
Successful transfer between different modalities (e.g. natural language to images) can be interpreted as less of a pretraining of transferable representations but instead transfer of the general \emph{computational structure} inherent in language to other tasks~\cite{Lu2021pretrained}.
Indeed, \citeauthor{Lu2021pretrained} find that finetuning only the input and output layers of a fully trained NLP transformer model, \emph{frozen} pretrained transformers~(FPTs), matches or outperforms training from scratch of the same model, across a variety of tasks in different modalities. 

In this paper, we report that the performance gains of FPTs disappear under fair tuning of the learning rate and one of the main claims of the paper no longer holds. 
FTPs perform better than random frozen models but are significantly worse than training from scratch on all four tasks studied.
Concretely, with a lower learning rate, the ordering of the performance between the unfrozen and frozen variants is inverted for all tasks (see~Figure~\ref{fig:lrsweep_cifar10_lra_annotated}).
The impact of hyperparamters on the empirical results in ML papers has been the subject of intense debates. 
It is not uncommon for careful finetuning to change the outcome of a paper or even undo years of ``progress'' in the field.
For example, \citet{melis2018on} found that when fairly optimizing for hyperparamters, vanilla LSTM match more sophisticated recently introduced RNN variants across a variety of NLP tasks.

That said, interestingly, we find that when \emph{not} frozen, Transformers do provide gains through transfer from text to other modalities. 
In particular in the challenging CIFAR10-LRA task \cite{tay2021long}, which consists of sequentialized CIFAR images, finetuning the entire pretrained model outperforms training from scratch by a large margin. On MNIST \cite{lecun-mnisthandwrittendigit-2010}, the gap is small but significant and on the other two tasks finetuning the pretrained model matches the performance of training from scratch.
This opens up exciting avenues for future research and we hope that our work will help the community avoid some potential pitfalls around hyperparemter tuning in the pursuit of this work, ensuring that the findings will stand the test of time.

\section{Problem Setting and Background}
\label{sec:background}

The recent work from \citet{Lu2021pretrained} investigates the capability of transformers, pretrained on natural language, to generalize to other modalities with minimal finetuning. They limit the finetuning by freezing the majority of the weights in the residual layers, and report that this Frozen Pretrained Transformer (FPT) architecture achieves comparable or better results than transformers trained from scratch across their chosen tasks.

\citeauthor{Lu2021pretrained} consider classification tasks across a range of modalities, including bit manipulation~\cite{pmlr-v80-miconi18a}, equation evaluation~\cite{tay2021long,nangia-bowman-2018-listops}, image classification~\cite{cifar10-dataset} and protein classification~\cite{NEURIPS2019_37f65c06,scope-fox-2013,deepsf-hou-2018}.  
The input is chunked into tokens for processing with the GPT2 architecture, with the tokens for the vision tasks being a flattened representation of a 4 by 4 pixel patch. The authors also include a more challenging version of the CIFAR10 task, CIFAR-LRA \cite{tay2021long} where the patches are a single pixel, resulting in longer sequences.

FPTs as proposed by \citeauthor{Lu2021pretrained} have the feedforward and multi-head attention frozen in each of the residual blocks. Only the input and output layers, layer norm parameters, and positional embeddings are finetuned. The authors compare this performance with a Frozen Random transformer and an Unfrozen variant.  
For the Unfrozen variant they report numbers from different architectures for each task we consider. For CIFAR10-LRA and ListOps they report numbers from a vanilla Transformer with tuned hyperparameters as provided in \citet{tay2021long}. 
For MNIST and CIFAR10 they report results from GPT2, with CIFAR10 using a 3 layer model due to instability in training the full sized version.

The authors report training on a single learning rate ($1 \times 10^{-3}$) for all tasks except Homology, and appear to report the results from a single seed per variation.  The released code\footnote{https://github.com/kzl/universal-computation} along with the paper does not use a validation set and they report the test accuracies from a held-out test set.

\section{Methods}
\label{methodology}
Training of deep neural networks can be highly sensitive to the learning rates used for optimizing the network~\cite{choi2020on}. Therefore, a natural question  is to ask whether the results reported in~\citet{Lu2021pretrained} have been impacted by the choice of using a fixed learning rate.  
To investigate this, we rerun the experiments of~\citeauthor{Lu2021pretrained} while broadly sweeping the learning rate. 
As we will see later, any given \emph{fixed} learning rate greatly changes the results. 

To investigate the effects of pretraining and freezing across tasks from different modalities, we evaluate on four of the tasks explored by~\citeauthor{Lu2021pretrained}: ListOps, MNIST, CIFAR10 and CIFAR10-LRA.  We do not replicate the Bit tasks because the transformers were able to perfectly solve them in the original work. The Homology task is not supported in the released codebase so it would be more difficult to ensure an accurate reproduction of their experimental setting.   
We evaluate the performance on the base GPT-2 model, at 12 layers. As in their work, we experiment with using transformer models pretrained on natural language, and with freezing the self-attention and feedforward layers finetuning only the input and output layers, the layer norm and the positional embeddings. Specifically, we consider:

\textbf{Frozen Pretrained}: The Frozen Pretrained Transformer introduced by ~\citeauthor{Lu2021pretrained}

\textbf{Frozen Random}: The transformer is randomly initialized and the self-attention and feedforward layers are frozen before finetuning.

\textbf{Unfrozen Pretrained}: The transformer is initialized with a pretrained language model and finetuned without freezing any layers.

\textbf{Unfrozen Random}: The transformer is randomly initialized and finetuned without freezing any layers.

For each of these settings and tasks, we train using the Adam optimizer \cite{kingma2015adam} and sweep the learning rate logarithmically from $1\times10^{-6}$ to $0.01$. We use a batch size of eight and train up to a predefined maximum number of gradient steps (see Appendix \ref{sample-table} for details). We determine the training step for early stopping based on the performance on the validation set to obtain the best model from each run and to identify the best learning rate across the sweep. For each setting, we repeat the experiments with three seeds and report the mean test accuracy along with the standard error of the mean as measured on the held-out test set.

For the ListOps task the validation split is provided as part of the dataset released with the Long Range Arena~\cite{tay2021long}, but for the CIFAR10 and MNIST datasets we create our own validation set.  The 50K image train split in the CIFAR10 dataset is further subdivided into 5 training batches of 10K images, each containing a balanced number of samples from each class.  We select one of these batches to be the validation set and train on the other 40K images.  For MNIST, we split the 60K image training set into a 50K image training split and a 10K image validation set by randomly selecting 1000 images from each class.

For the ListOps task we note that the codebase released by \citeauthor{Lu2021pretrained} is affected by issues only recently fixed in the Long Range Arena codebase.\footnote{https://github.com/google-research/long-range-arena} Specifically, the ListOps dataset includes sequences ranging from 500 to 2000 tokens, but the dataset tokenization utility truncated all sequences to 512 tokens. In addition, the ``close'' parenthesis for the list of operations was not tokenized due to not being an alphanumeric character.  Between these two issues it was impossible to solve the long sequences of operations provided by the dataset.  We resolved these issues in the dataset tokenization utility and adapted the architecture choices accordingly, by increasing the context length and the number of positions used by the transformer architecture to 2000. This change is possible because in all settings we fine-tune the positional embeddings.  The additional tokenized character increased the token dimension for the ListOps task to 16.

\citeauthor{Lu2021pretrained} report that the model capacity impacts the performance of each of the settings, with increased model capacity hurting the performance of the Unfrozen variants and helping the performance of the Frozen variants, resulting in some of the Unfrozen results being reported for models with 3 layers instead of the full 12 layer GPT2. To evaluate the impact of model capacity and to provide a datapoint between the two model sizes, we also test using a pretrained DistilGPT2 which is a 6 layer transformer distilled from the full sized pretrained GPT2 model.

\section{Results and Discussion}
\label{sec:results}
While at a high level, our work confirms the finding from \citeauthor{Lu2021pretrained} that transfer from NLP tasks to other modalities is indeed possible through finetuning, our results contradict theirs regarding which parts of the model should be fine-tuned. 
Our main finding is that while the \emph{Unfrozen} Pretrained variant matches or outperforms all other settings across all tasks explored, the \emph{Frozen} variants often greatly lag in performance comparatively, in direct contradiction to their findings.  
Table \ref{main-table} compares between the different settings for each of the tasks across 3 seeds. For each task, the Frozen Pretrained setting outperforms the Frozen Random setting.  However, in contrast to their results, the Unfrozen variants always outperform the Frozen variants and by a large margin for all the tasks except for MNIST.

\begin{table*}[t]
\caption{Comparison of test accuracy across initialization and finetuning methods for the GPT2 architecture.}
\label{main-table}
\vskip 0.15in
\begin{center}
\begin{small}
\begin{sc}
\begin{tabular}{l|c|c|c|c}
\toprule
 & ListOps & MNIST & CIFAR10 & CIFAR10-LRA \\
 \midrule
Frozen Random       & 38.9 $\pm$  0.3 & 98.0 $\pm$  0.0 & 61.8 $\pm$  0.2 & 44.2 $\pm$  0.3 \\
Frozen Pretrained   & 46.1 $\pm$  0.3 & 98.5 $\pm$  0.1 & 66.3 $\pm$  0.0 & 54.7 $\pm$  1.4 \\
Unfrozen Random     & \textbf{57.6 $\pm$  0.8} & 98.7 $\pm$  0.0 & \textbf{77.8 $\pm$  0.2} & 62.0 $\pm$  0.7 \\
Unfrozen Pretrained & \textbf{56.3 $\pm$  0.9} & \textbf{99.0 $\pm$  0.0} & \textbf{77.7 $\pm$  0.1} & \textbf{67.8 $\pm$  0.3} \\
\bottomrule
\end{tabular}
\end{sc}
\end{small}
\end{center}
\vskip -0.1in
\end{table*}

The differences between these results and the ones obtained and reported by \citeauthor{Lu2021pretrained} in their Section 3.1, 3.2 and 3.11 can be explained by investigating the test accuracy across the learning rate sweep, shown in Figure \ref{fig:lrsweep_cifar10_lra_annotated} for the CIFAR10-LRA task.
Notably, the learning rate impacts not only the absolute performance but also the ordering between the settings.
The learning rate reported by \citeauthor{Lu2021pretrained}, $1 \times 10^{-3}$, is marked with a vertical dashed red line.
Since $1 \times 10^{-3}$ is just before a precipitous drop in the performance of the unfrozen transformers, had the authors picked a lower learning rate they would have arrived at very different conclusions.

When repeating this analysis for the ListOps task, in Figure \ref{fig:mainresults}, we see an even greater dependence on the learning rate. Each of the LRs evaluated between $1 \times 10^-5$ and $1 \times 10^-3$ results in different orderings and, hence, conclusions about the optimal architecture variant.  See Appendix \ref{appendix_lr_sweep} for plots for MNIST and CIFAR10 tasks which uphold these findings.

The key shared finding between \citeauthor{Lu2021pretrained} and our work is the benefit of finetuning from a model pretrained on natural language, even for tasks of different modalities. 
In their Section 3.2, \citeauthor{Lu2021pretrained} find that the Frozen Pretrained transformer is superior to a Frozen Random variant.
We verify that pretraining improves performance across all tasks for the Frozen transformers, and in addition find that for some tasks pretraining provides benefits for finetuning the Unfrozen variants.
For the CIFAR10-LRA task, the Unfrozen Pretrained variant outperforms all other variants by 4.8\%, and on MNIST the Unfrozen Pretrained variant outperforms the rest by a small margin.
This benefit from pretraining on some tasks, paired with matched performance on the rest, suggests that it may be expedient to run initial experiments in new settings by finetuning from a natural language pretrained model.
However the varying success of pretraining across tasks raises an open question, for future work, about which qualities of a task lead to benefits from pretraining.

In their Section 3.4, \citeauthor{Lu2021pretrained} compare the computation efficiency between the Frozen Random and Frozen Pretrained variants by reporting the number of gradient steps to convergence. When sweeping the learning rate, we see that the final performance of the Frozen variants is much lower than the Unfrozen variants. Thus, we instead compare computational efficiency by reporting the number of gradient steps to match the performance of the Frozen Pretrained variant in Appendix \ref{appendix_compute_efficiency}. The Frozen Random variant does not match performance in any of the tasks, verifying the authors' assertion that pretraining improves the computational efficiency of the Frozen variants. However, for all tasks the Unfrozen variants require fewer gradient steps. For all but ListOps the Unfrozen Pretrained variant requires the least gradient steps, demonstrating that in some tasks pretraining helps not only final performance but also computational efficiency.

In their Section 3.6, \citeauthor{Lu2021pretrained} report that the Frozen Pretrained models underfit the CIFAR10 task.  We validate these findings (in Appendix \ref{appendix_underfitting}) but argue that underfitting is likely what leads to the poor comparative performance of the Frozen variants in our experiments. In addition, while the Frozen variants underfit on the MNIST and CIFAR10 tasks, the Frozen Pretrained variant has the largest train/test gap of all settings on ListOps, invalidating the hypothesis that the Frozen variants always underfit the data.

In their Section 3.7, \citeauthor{Lu2021pretrained} report that increasing the model capacity of the Frozen Pretrained setting improved the performance on the CIFAR10 task, suggesting an easy way to achieve performance gains. We report results from similar experiments, with the addition of the learning rate sweep, in Appendix \ref{appendix_scaling}, confirming their finding of performance gains from increased capacity for the Frozen Pretrained variant on CIFAR10. Our results also verify that these claims hold for the CIFAR10-LRA task and but show that increasing model capacity also benefits the Unfrozen Pretrained variant.  
Interestingly, the increase in model capacity improves the Unfrozen Pretrained setting across all tasks whereas it only improves the Unfrozen Random setting on CIFAR10-LRA.

In their Section 3.11, \citeauthor{Lu2021pretrained} describe the impact of unfreezing part or all of the pretrained model.  
They report that unfreezing both the feedforward layers and the attention heads is detrimental to performance.  This experiment corresponds to our Unfrozen Pretrained variant which we find outperforms all other variants when the learning rate is properly swept, contrary to their findings.

\section{Conclusion}
\label{conclusion}
Transformer architectures pretrained on natural language texts are some of the most successful models in NLP in recent years. Therefore, investigating how they can best be used as a starting point for solving tasks in other modalities is an important research direction. In our work, we show that, across a variety of tasks, the best results are obtained when finetuning all of the weights of a pretrained model. These results directly contradict prior work which concluded that freezing most of the model leads to superior performance. We demonstrate that these prior conclusions were an artefact of using a specific, fixed learning rate, and hope that our work will help pave the way for robust investigations into cross-modal transfer in the future. 

\clearpage

\section{Acknowledgements}
The authors wish to thank Edward Grefenstette, Hengyuan Hu and Patrick Lewis for many helpful discussions during the initial stages of this project.

\bibliographystyle{icml2021}
\bibliography{main.bib}

\begin{thebibliography}{27}
\providecommand{\natexlab}[1]{#1}
\providecommand{\url}[1]{\texttt{#1}}
\expandafter\ifx\csname urlstyle\endcsname\relax
  \providecommand{\doi}[1]{doi: #1}\else
  \providecommand{\doi}{doi: \begingroup \urlstyle{rm}\Url}\fi

\bibitem[Brown et~al.(2020)Brown, Mann, Ryder, Subbiah, Kaplan, Dhariwal,
  Neelakantan, Shyam, Sastry, Askell, Agarwal, Herbert-Voss, Krueger, Henighan,
  Child, Ramesh, Ziegler, Wu, Winter, Hesse, Chen, Sigler, Litwin, Gray, Chess,
  Clark, Berner, McCandlish, Radford, Sutskever, and
  Amodei]{NEURIPS2020_1457c0d6}
Brown, T., Mann, B., Ryder, N., Subbiah, M., Kaplan, J.~D., Dhariwal, P.,
  Neelakantan, A., Shyam, P., Sastry, G., Askell, A., Agarwal, S.,
  Herbert-Voss, A., Krueger, G., Henighan, T., Child, R., Ramesh, A., Ziegler,
  D., Wu, J., Winter, C., Hesse, C., Chen, M., Sigler, E., Litwin, M., Gray,
  S., Chess, B., Clark, J., Berner, C., McCandlish, S., Radford, A., Sutskever,
  I., and Amodei, D.
\newblock Language models are few-shot learners.
\newblock In \emph{Advances in Neural Information Processing Systems},
  volume~33, pp.\  1877--1901. Curran Associates, Inc., 2020.

\bibitem[Chelba et~al.(2014)Chelba, Mikolov, Schuster, Ge, Brants, Koehn, and
  Robinson]{chelba2014one}
Chelba, C., Mikolov, T., Schuster, M., Ge, Q., Brants, T., Koehn, P., and
  Robinson, T.
\newblock One billion word benchmark for measuring progress in statistical
  language modeling.
\newblock In \emph{Fifteenth Annual Conference of the International Speech
  Communication Association}, 2014.

\bibitem[Choi et~al.(2020)Choi, Shallue, Nado, Lee, Maddison, and
  Dahl]{choi2020on}
Choi, D., Shallue, C.~J., Nado, Z., Lee, J., Maddison, C.~J., and Dahl, G.~E.
\newblock On empirical comparisons of optimizers for deep learning.
\newblock In \emph{International Conference on Learning Representations}, 2020.

\bibitem[Devlin et~al.(2019)Devlin, Chang, Lee, and
  Toutanova]{devlin-etal-2019-bert}
Devlin, J., Chang, M.-W., Lee, K., and Toutanova, K.
\newblock {BERT}: Pre-training of deep bidirectional transformers for language
  understanding.
\newblock In \emph{Proceedings of the 2019 Conference of the North {A}merican
  Chapter of the Association for Computational Linguistics: Human Language
  Technologies, Volume 1 (Long and Short Papers)}, pp.\  4171--4186,
  Minneapolis, Minnesota, June 2019. Association for Computational Linguistics.
\newblock \doi{10.18653/v1/N19-1423}.

\bibitem[Fox et~al.(2013)Fox, Brenner, and Chandonia]{scope-fox-2013}
Fox, N.~K., Brenner, S.~E., and Chandonia, J.-M.
\newblock Scope: Structural classification of proteins--extended, integrating
  scop and astral data and classification of new structures.
\newblock In \emph{Nucleic Acids Research}, pp.\  D304–D309, 2013.
\newblock \doi{10.1093/nar/gkt1240}.

\bibitem[Hou et~al.(2018)Hou, Adhikari, and Cheng]{deepsf-hou-2018}
Hou, J., Adhikari, B., and Cheng, J.
\newblock Deepsf: Deep convolutional neural network for mapping protein
  sequences to folds.
\newblock In \emph{Bioinformatics}, pp.\  1295--1303, 08 2018.
\newblock \doi{doi:10.1093/bioinformatics/btx780}.

\bibitem[Kingma \& Ba(2015)Kingma and Ba]{kingma2015adam}
Kingma, D.~P. and Ba, J.
\newblock Adam: A method for stochastic optimization.
\newblock In \emph{International Conference on Learning Representations}, 2015.

\bibitem[Krizhevsky(2012)]{cifar10-dataset}
Krizhevsky, A.
\newblock Learning multiple layers of features from tiny images.
\newblock \emph{University of Toronto}, 05 2012.

\bibitem[LeCun \& Cortes(2010)LeCun and
  Cortes]{lecun-mnisthandwrittendigit-2010}
LeCun, Y. and Cortes, C.
\newblock {MNIST} handwritten digit database.
\newblock 2010.
\newblock URL \url{http://yann.lecun.com/exdb/mnist/}.

\bibitem[Lu et~al.(2021)Lu, A.~Grover, Abbeel, and Mordatch]{Lu2021pretrained}
Lu, K., A.~Grover, A., Abbeel, P., and Mordatch, I.
\newblock Pretrained transformers as universal computation engines.
\newblock \emph{arXiv:2103.05247}, 2021.

\bibitem[Maas et~al.(2011)Maas, Daly, Pham, Huang, Ng, and
  Potts]{maas-etal-2011-learning}
Maas, A.~L., Daly, R.~E., Pham, P.~T., Huang, D., Ng, A.~Y., and Potts, C.
\newblock Learning word vectors for sentiment analysis.
\newblock In \emph{Proceedings of the 49th Annual Meeting of the Association
  for Computational Linguistics: Human Language Technologies}, pp.\  142--150,
  Portland, Oregon, USA, June 2011. Association for Computational Linguistics.

\bibitem[Melis et~al.(2018)Melis, Dyer, and Blunsom]{melis2018on}
Melis, G., Dyer, C., and Blunsom, P.
\newblock On the state of the art of evaluation in neural language models.
\newblock In \emph{International Conference on Learning Representations}, 2018.

\bibitem[Miconi et~al.(2018)Miconi, Stanley, and Clune]{pmlr-v80-miconi18a}
Miconi, T., Stanley, K., and Clune, J.
\newblock Differentiable plasticity: training plastic neural networks with
  backpropagation.
\newblock In Dy, J. and Krause, A. (eds.), \emph{Proceedings of the 35th
  International Conference on Machine Learning}, volume~80 of \emph{Proceedings
  of Machine Learning Research}, pp.\  3559--3568. PMLR, 10--15 Jul 2018.

\bibitem[Mikolov et~al.(2013)Mikolov, Sutskever, Chen, Corrado, and
  Dean]{NIPS2013_9aa42b31}
Mikolov, T., Sutskever, I., Chen, K., Corrado, G.~S., and Dean, J.
\newblock Distributed representations of words and phrases and their
  compositionality.
\newblock In Burges, C. J.~C., Bottou, L., Welling, M., Ghahramani, Z., and
  Weinberger, K.~Q. (eds.), \emph{Advances in Neural Information Processing
  Systems}, volume~26. Curran Associates, Inc., 2013.

\bibitem[Nangia \& Bowman(2018)Nangia and Bowman]{nangia-bowman-2018-listops}
Nangia, N. and Bowman, S.
\newblock {L}ist{O}ps: A diagnostic dataset for latent tree learning.
\newblock In \emph{Proceedings of the 2018 Conference of the North {A}merican
  Chapter of the Association for Computational Linguistics: Student Research
  Workshop}, pp.\  92--99, New Orleans, Louisiana, USA, June 2018. Association
  for Computational Linguistics.
\newblock \doi{10.18653/v1/N18-4013}.

\bibitem[Pennington et~al.(2014)Pennington, Socher, and
  Manning]{pennington-etal-2014-glove}
Pennington, J., Socher, R., and Manning, C.
\newblock {G}lo{V}e: Global vectors for word representation.
\newblock In \emph{Proceedings of the 2014 Conference on Empirical Methods in
  Natural Language Processing ({EMNLP})}, pp.\  1532--1543, Doha, Qatar,
  October 2014. Association for Computational Linguistics.
\newblock \doi{10.3115/v1/D14-1162}.

\bibitem[Petroni et~al.(2019)Petroni, Rockt{\"a}schel, Riedel, Lewis, Bakhtin,
  Wu, and Miller]{petroni-etal-2019-language}
Petroni, F., Rockt{\"a}schel, T., Riedel, S., Lewis, P., Bakhtin, A., Wu, Y.,
  and Miller, A.
\newblock Language models as knowledge bases?
\newblock In \emph{Proceedings of the 2019 Conference on Empirical Methods in
  Natural Language Processing and the 9th International Joint Conference on
  Natural Language Processing (EMNLP-IJCNLP)}, pp.\  2463--2473, Hong Kong,
  China, November 2019. Association for Computational Linguistics.
\newblock \doi{10.18653/v1/D19-1250}.

\bibitem[Petroni et~al.(2021)Petroni, Piktus, Fan, Lewis, Yazdani, De~Cao,
  Thorne, Jernite, Karpukhin, Maillard, Plachouras, Rockt{\"a}schel, and
  Riedel]{petroni-etal-2021-kilt}
Petroni, F., Piktus, A., Fan, A., Lewis, P., Yazdani, M., De~Cao, N., Thorne,
  J., Jernite, Y., Karpukhin, V., Maillard, J., Plachouras, V.,
  Rockt{\"a}schel, T., and Riedel, S.
\newblock {KILT}: a benchmark for knowledge intensive language tasks.
\newblock In \emph{Proceedings of the 2021 Conference of the North American
  Chapter of the Association for Computational Linguistics: Human Language
  Technologies}, pp.\  2523--2544, 2021.

\bibitem[Radford et~al.(2019)Radford, Wu, Child, Luan, Amodei, and
  Sutskever]{radford2019language}
Radford, A., Wu, J., Child, R., Luan, D., Amodei, D., and Sutskever, I.
\newblock Language models are unsupervised multitask learners.
\newblock \emph{OpenAI blog}, 1\penalty0 (8):\penalty0 9, 2019.

\bibitem[Rao et~al.(2019)Rao, Bhattacharya, Thomas, Duan, Chen, Canny, Abbeel,
  and Song]{NEURIPS2019_37f65c06}
Rao, R., Bhattacharya, N., Thomas, N., Duan, Y., Chen, P., Canny, J., Abbeel,
  P., and Song, Y.
\newblock Evaluating protein transfer learning with tape.
\newblock In Wallach, H., Larochelle, H., Beygelzimer, A., d\textquotesingle
  Alch\'{e}-Buc, F., Fox, E., and Garnett, R. (eds.), \emph{Advances in Neural
  Information Processing Systems}, volume~32. Curran Associates, Inc., 2019.

\bibitem[Tay et~al.(2021)Tay, Dehghani, Abnar, Shen, Bahri, Pham, Rao, Yang,
  Ruder, and Metzler]{tay2021long}
Tay, Y., Dehghani, M., Abnar, S., Shen, Y., Bahri, D., Pham, P., Rao, J., Yang,
  L., Ruder, S., and Metzler, D.
\newblock Long range arena : A benchmark for efficient transformers.
\newblock In \emph{International Conference on Learning Representations}, 2021.

\bibitem[Vaswani et~al.(2017)Vaswani, Shazeer, Parmar, Uszkoreit, Jones, Gomez,
  Kaiser, and Polosukhin]{10.5555/3295222.3295349}
Vaswani, A., Shazeer, N., Parmar, N., Uszkoreit, J., Jones, L., Gomez, A.~N.,
  Kaiser, u., and Polosukhin, I.
\newblock Attention is all you need.
\newblock In \emph{Proceedings of the 31st International Conference on Neural
  Information Processing Systems}, NIPS'17, pp.\  6000–6010, Red Hook, NY,
  USA, 2017. Curran Associates Inc.
\newblock ISBN 9781510860964.

\bibitem[Wang et~al.(2018)Wang, Singh, Michael, Hill, Levy, and
  Bowman]{wang-etal-2018-glue}
Wang, A., Singh, A., Michael, J., Hill, F., Levy, O., and Bowman, S.
\newblock {GLUE}: A multi-task benchmark and analysis platform for natural
  language understanding.
\newblock In \emph{Proceedings of the 2018 {EMNLP} Workshop {B}lackbox{NLP}:
  Analyzing and Interpreting Neural Networks for {NLP}}, pp.\  353--355,
  Brussels, Belgium, November 2018. Association for Computational Linguistics.
\newblock \doi{10.18653/v1/W18-5446}.

\bibitem[Wang et~al.(2019)Wang, Pruksachatkun, Nangia, Singh, Michael, Hill,
  Levy, and Bowman]{NEURIPS2019_4496bf24}
Wang, A., Pruksachatkun, Y., Nangia, N., Singh, A., Michael, J., Hill, F.,
  Levy, O., and Bowman, S.
\newblock Superglue: A stickier benchmark for general-purpose language
  understanding systems.
\newblock In Wallach, H., Larochelle, H., Beygelzimer, A., d\textquotesingle
  Alch\'{e}-Buc, F., Fox, E., and Garnett, R. (eds.), \emph{Advances in Neural
  Information Processing Systems}, volume~32. Curran Associates, Inc., 2019.

\bibitem[Wolf et~al.(2020)Wolf, Debut, Sanh, Chaumond, Delangue, Moi, Cistac,
  Rault, Louf, Funtowicz, Davison, Shleifer, von Platen, Ma, Jernite, Plu, Xu,
  Le~Scao, Gugger, Drame, Lhoest, and Rush]{wolf-etal-2020-transformers}
Wolf, T., Debut, L., Sanh, V., Chaumond, J., Delangue, C., Moi, A., Cistac, P.,
  Rault, T., Louf, R., Funtowicz, M., Davison, J., Shleifer, S., von Platen,
  P., Ma, C., Jernite, Y., Plu, J., Xu, C., Le~Scao, T., Gugger, S., Drame, M.,
  Lhoest, Q., and Rush, A.
\newblock Transformers: State-of-the-art natural language processing.
\newblock In \emph{Proceedings of the 2020 Conference on Empirical Methods in
  Natural Language Processing: System Demonstrations}, pp.\  38--45, Online,
  October 2020. Association for Computational Linguistics.
\newblock \doi{10.18653/v1/2020.emnlp-demos.6}.

\bibitem[Zhang et~al.(2015)Zhang, Zhao, and LeCun]{NIPS2015_250cf8b5}
Zhang, X., Zhao, J., and LeCun, Y.
\newblock Character-level convolutional networks for text classification.
\newblock In Cortes, C., Lawrence, N., Lee, D., Sugiyama, M., and Garnett, R.
  (eds.), \emph{Advances in Neural Information Processing Systems}, volume~28.
  Curran Associates, Inc., 2015.

\bibitem[Zhu et~al.(2015)Zhu, Kiros, Zemel, Salakhutdinov, Urtasun, Torralba,
  and Fidler]{10.1109/ICCV.2015.11}
Zhu, Y., Kiros, R., Zemel, R., Salakhutdinov, R., Urtasun, R., Torralba, A.,
  and Fidler, S.
\newblock Aligning books and movies: Towards story-like visual explanations by
  watching movies and reading books.
\newblock In \emph{Proceedings of the 2015 IEEE International Conference on
  Computer Vision (ICCV)}, ICCV '15, pp.\  19–27, USA, 2015. IEEE Computer
  Society.
\newblock ISBN 9781467383912.
\newblock \doi{10.1109/ICCV.2015.11}.

\end{thebibliography}
\newpage

\newpage
\clearpage 
\appendix
\onecolumn

\section{Learning Rate Sweeps}
\label{appendix_lr_sweep}
We report the impact of the learning rate on the test accuracy for all tasks, MNIST, ListOps and CIFAR10 and CIFAR10-LRA. In all cases, the learning rate reported by \citet{Lu2021pretrained} is marked with a dashed red line and it is clear that reducing the reported learning rate would invert the findings.

\begin{figure}[h]
	\vspace*{0.5em}
    \centerline{\includegraphics[width=12cm]{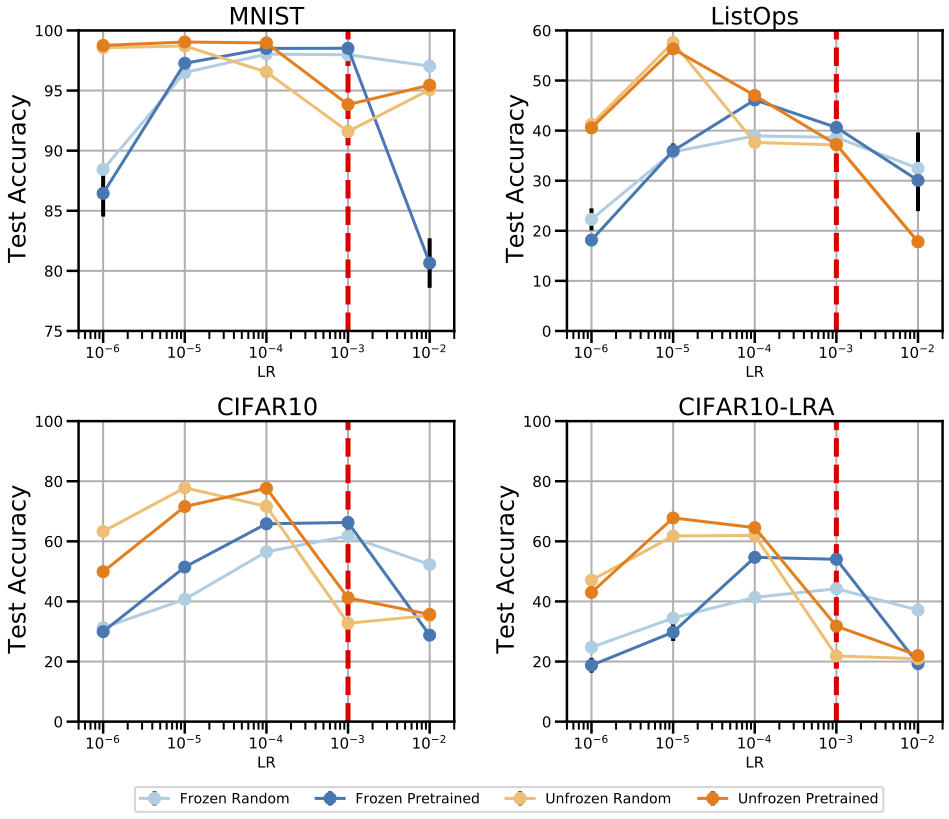}}
    \caption{Test accuracy of all tasks across the learning rate sweep, error bars across 3 seeds.}
    \label{fig:lrsweep_all_tasks}
\end{figure}

\section{Computational Efficiency}
\label{appendix_compute_efficiency}
We investigate the impact of pretraining and freezing on the computational efficiency of the architectures. Because the final performance varies dramatically between variants we compare the number of gradient steps necessary to reach the best test accuracy of the Frozen Pretrained variant.  We find that the Frozen Random variant is never able to match this performance.  The Unfrozen settings require fewer gradient steps to match performance across all tasks, with pretraining generally improving computational efficiency in three out of four tasks.

\begin{table*}[h]
\caption{The impact of pretraining on compute efficiency, comparing the number of gradient steps, per variant, to match the reported best mean test accuracy of the Frozen Pretrained variant.}
\label{gradient-steps-table}
\vskip 0.15in
\begin{center}
\begin{small}
\begin{sc}
\begin{tabular}{l|c|c|c|c}
\toprule
 & ListOps & MNIST & CIFAR10 & CIFAR10-LRA \\
 \midrule
Unfrozen Random & \bm{$1.0\times 10^5$} & $1.1\times 10^5$ & $7.0\times 10^4$ & $1.8\times 10^5$ \\
Frozen Random &  - &  - &  - &  - \\
Unfrozen Pretrained & $1.9\times 10^5$ & \bm{$4.6\times 10^4$} & \bm{$4.3\times 10^4$} & \bm{$5.3\times 10^4$} \\
Frozen Pretrained & $1.9\times 10^5$ & $2.4\times 10^5$ & $3.8\times 10^5$ & $2.4\times 10^5$ \\
\bottomrule
\end{tabular}
\end{sc}
\end{small}
\end{center}
\vskip -0.1in
\end{table*}

\section{Underfitting versus Overfitting}
\label{appendix_underfitting}
We investigate the extent to which each of the variants is able to fit the data by reporting the train and test accuracy at the threshold where we ended training, specified in Table \ref{sample-table} for each task.  We find that across the MNIST and CIFAR10 tasks the Frozen variants underfit the data.  However, this trend does not hold for the ListOps task where the Frozen Pretrained setting has the largest train/test gap.

\begin{table*}[h]
\caption{Train versus Test accuracy at the maximum number of gradient steps taken for each task as listed in Table \ref{sample-table}.}
\label{train-test-table}
\vskip 0.15in
\begin{center}
\begin{small}
\begin{sc}
\begin{tabular}{l|c|c|c|c|c}
\toprule
 & & ListOps & MNIST & CIFAR10 & CIFAR10-LRA \\
 \midrule
Unfrozen Random & Train & 58.1 &  99.9 &  96.4 &  74.1 \\
                & Test &  57.5 &  98.6 &  77.7 &  61.7 \\
                & Diff &   0.6 &   1.3 &  18.7 &  12.4 \\
                \hline
Frozen Random & Train &  39.7 &  98.8 &  62.9 &  44.9 \\
 & Test &  39.2 &  98.0 &  61.1 &  43.8 \\
 & Diff &   0.5 &   0.8 &   1.9 &   1.1 \\
 \hline
Unfrozen Pretrained & Train & 57.8 & 100.0 &  98.7 &  91.0 \\
 & Test &  55.7 &  99.0 &  77.1 &  67.0 \\
 & Diff &   2.1 &   1.0 &  21.7 &  24.0 \\
 \hline
Frozen Pretrained & Train & 52.2 &  99.3 &  67.8 &  55.6 \\
 & Test & 46.4 &  98.5 &  65.5 &  53.4 \\
 & Diff &   5.8 &   0.8 &   2.3 &   2.2 \\
\bottomrule
\end{tabular}
\end{sc}
\end{small}
\end{center}
\vskip -0.1in
\end{table*}

\section{Scaling Model Capacity}
\label{appendix_scaling}
We investigate the impact of scaling the model capacity across three of the tasks, MNIST, CIFAR10 and CIFAR10-LRA.  We compare the DistilGPT2 model at 6 layers against the GPT2 base model at 12 layers, both provided by the HuggingFace Transformers library \cite{wolf-etal-2020-transformers}.  Scaling has little or no impact on MNIST and the only variant to show improvement across all tasks with increased model capacity is the Unfrozen Pretrained setting.  The Frozen Pretrained setting also improves with model capacity on both CIFAR10 tasks.

\begin{table*}[h]
\caption{The impact of scaling the size of the transformers across three of the tasks, comparing the performance of the DistilGPT2 architecture with that of the GPT2 architecture.}
\label{scaling-table}
\vskip 0.15in
\begin{center}
\begin{small}
\begin{sc}
\begin{tabular}{l|l|c|c|c}
\toprule
 & & MNIST & CIFAR10 & CIFAR10-LRA \\
 \midrule

Frozen Random       & distilgpt2 & \textbf{98.0 $\pm$  0.1} & 60.1 $\pm$  0.1 & \textbf{45.0 $\pm$  0.1} \\
                    & gpt2 & \textbf{98.0 $\pm$  0.0} & \textbf{61.8 $\pm$  0.2} & 44.2 $\pm$  0.3 \\
\hline
Frozen Pretrained   & distilgpt2 & \textbf{98.5 $\pm$  0.1} & 65.2 $\pm$  0.5 & 51.1 $\pm$  0.4 \\
                    & gpt2 & \textbf{98.5 $\pm$  0.1} & \textbf{66.3 $\pm$  0.0} & \textbf{54.7 $\pm$  1.4} \\
\hline
Unfrozen Random     & distilgpt2 & \textbf{98.6 $\pm$  0.1} & \textbf{77.5 $\pm$  0.1} & 59.7 $\pm$  0.2 \\
                    & gpt2 & \textbf{98.7 $\pm$  0.0} & \textbf{77.8 $\pm$  0.2} & \textbf{62.0 $\pm$  0.7} \\
\hline
Unfrozen Pretrained & distilgpt2 & 98.9 $\pm$  0.0 & 76.8 $\pm$  0.1 & 65.5 $\pm$  0.5 \\
                    & gpt2 & \textbf{99.0 $\pm$  0.0} & \textbf{77.7 $\pm$  0.1} & \textbf{67.8 $\pm$  0.3} \\
\bottomrule
\end{tabular}
\end{sc}
\end{small}
\end{center}
\vskip -0.1in
\end{table*}

\section{Evaluation Setup}
We trained each of the tasks for a logarithmic sweep of learning rates, from $1 \times 10^{-6}$ to $1 \times 10^{-2}$. Each task was run for a fixed number of gradient steps, specified in Table \ref{sample-table}.  The validation accuracy was used to perform early stopping and to identify the model in each run to evaluate and the test accuracy from that model is reported.

\begin{table}[h]
\caption{Threshold number of gradient steps used to report test accuracy results, per task and model type.}
\label{sample-table}
\vskip 0.15in
\begin{center}
\begin{small}
\begin{sc}
\begin{tabular}{l|c|c}
\toprule
Task & Model Type & Number Gradient \\
     &           & Steps            \\
\midrule
ListOps    & GPT2&          $3\times10^5$ \\
MNIST      & GPT2&          $4\times10^5$ \\
CIFAR10    & GPT2&          $4\times10^5$ \\
CIFAR10 LRA& GPT2&          $3\times10^5$ \\
\bottomrule
\end{tabular}
\end{sc}
\end{small}
\end{center}
\vskip -0.1in
\end{table}

\end{document}